\documentclass[review]{elsarticle}

\usepackage{lineno,hyperref}
\modulolinenumbers[5]

\journal{Artificial Intelligence}

\usepackage{amsmath, amsfonts, bbm}
\usepackage{amsthm,amssymb}
\usepackage{array}

\newtheorem{defn}{Definition}
\newtheorem{example}{Example}
\newcolumntype{M}[1]{>{\centering\arraybackslash}m{#1}}

\usepackage{lipsum}

\newcommand\blfootnote[1]{%
  \begingroup
  \renewcommand\thefootnote{}\footnote{#1}%
  \addtocounter{footnote}{-1}%
  \endgroup
}

\begin{document}

\begin{frontmatter}

\title{Value of Information for Argumentation based Intelligence Analysis}

\author[mymainaddress]{Todd Robinson}
\ead{trobinson2@dstl.gov.uk}

\address[mymainaddress]{Defence Science and Technology Laboratory, Porton Down, Salisbury, United Kingdom, SP4 0JQ}

\begin{abstract}
Argumentation provides a representation of arguments and attacks between these arguments.  Argumentation can be used to represent a reasoning process over evidence to reach conclusions. Within such a reasoning process, understanding the value of information can improve the quality of decision making based on the output of the reasoning process. The value of an item of information is inherently dependent on the available evidence and the question being answered by the reasoning. In this paper we introduce a value of information on argument frameworks to identify the most valuable arguments within the finite set of arguments in the framework, and the arguments and attacks which could be added to change the output of an evaluation. We demonstrate the value of information within an argument framework representing an intelligence analysis in the maritime domain. Understanding the value of information in an intelligence analysis will allow analysts to balance the value against the costs and risks of collection, to effectively request further collection of intelligence to increase the confidence in the analysis of hypotheses.
\blfootnote{\copyright Crown copyright (2021), Dstl. This material is licensed under the terms of the Open Government Licence except where otherwise stated. To view this licence, visit http://www.nationalarchives.gov.uk/doc/open-government-licence/version/3 or write to the Information Policy Team, The National Archives, Kew, London TW9 4DU, or email: psi@nationalarchives.gov.uk}
\end{abstract}

\begin{keyword}
Argumentation \sep Probabilistic Argumentation \sep Value of Information \sep Intelligence Analysis
\end{keyword}

\end{frontmatter}


\section{Introduction}

Argumentation is the study of logical reasoning over evidence to reach conclusions. Argumentation is an inter-disciplinary area, with contributions from legal theory, linguistics, dialogics, mathematics and computer science. In this paper we concentrate on argumentation theory in the mathematical sense, where arguments are represented as a logical model which can be evaluated. Most argumentation theory is underpinned by Dung argument frameworks \cite{dung1995acceptability}, a representation of arguments and attacks between these arguments. Dung frameworks can be evaluated according to some semantics, to identify subsets of arguments which can be accepted together. Argumentation theory has also been extended to probabilistic argumentation to represent uncertain arguments and attacks. 

Understanding the value of information can improve the quality of a decision making process. The value of an item of information is inherently dependent on the available evidence and the question being answered. The value of information to the decision making process has not previously been considered in argument frameworks. In this paper we introduce a value of information on Dung and probabilistic argument frameworks. Within an argument framework, understanding the contribution of an argument towards a conclusion can help to identify the central arguments in an analysis and therefore possible weaknesses. Similarly we can consider the value of adding arguments and attacks to a framework in order to reinforce or change the conclusion of an argument map. In this paper we consider value of information for argumentation in intelligence analysis but the concepts developed also have applications to other domains. In the legal domain, within a legal argument represented in argumentation, consideration of the value of information would help the Defence and Prosecution identify the most valuable arguments to attack in order to get the result they desire. 

To perform an intelligence analysis, an analyst will begin with a body of evidence and will apply sensemaking techniques to build and evaluate hypotheses. This process can be represented in argumentation through the inferences and conflicts identified between evidence and hypotheses. Analysts also use structured analytical techniques \cite{uk2013quick}, many of which can be represented in argumentation, for example \cite{murukannaiah2015resolving}.

Intelligence analysis is not a passive process. Analysts can request collection of further evidence to improve their analysis. The collection of evidence can be expensive and may pose a risk to assets. It is therefore often not possible to collect all the evidence required to make a conclusion with full confidence. The value of making a collection to improve the confidence in the analysis outputs must be balanced against the cost and risk of making the collection. Currently, this judgement is made using analyst experience but with intelligence analyses represented in argumentation, the value of information will help identify the evidence which could be collected. 

Intelligence analysts are taught to be aware of the biases which are part of human reasoning, such as confirmation bias. Confirmation bias \cite{nickerson1998confirmation} is the tendency to select information which reinforces prior beliefs. Confirmation bias is one of the key biases which requires consideration for the value of information problem, as many approaches to value of information encourage increasing the certainty of the most likely hypotheses (i.e. by reducing entropy). Instead we consider an argument to be valuable if it changes the conclusion of an analysis, in order to minimise confirmation bias.

Section \ref{related} presents related work in the fields of argumentation, probabilistic argumentation and value of information. In section \ref{background}, we formally define argumentation theory in terms of Dung argument frameworks and probabilistic argument frameworks. In section \ref{VoI} we introduce value of information functions for arguments within an argument framework and for arguments and attacks which could be added to an argument framework. Section \ref{ApplicationIA} demonstrates how these value functions could be used in a maritime intelligence analysis problem represented in argumentation. Finally in section \ref{conclusions} we present conclusions and in section \ref{futurework} some recommendations for future work to develop these concepts further. 

\section{Related Work}\label{related}

\subsection{Argumentation}

The argumentation framework introduced by Dung \cite{dung1995acceptability} has made a major contribution to argumentation theory, and the semantics introduced by Dung underpin the evaluation of most argumentation systems, including ASPIC+ \cite{prakken2010abstract}. A good overview of argumentation theory is given in \cite{walton2009argumentation}, and semantics in \cite{baroni2011introduction}.

A number of representations of argumentation have been developed, including ASPIC+ \cite{prakken2010abstract}, description logic \cite{rahwan2011representing} and rationale \cite{ter2013critical}. Argument Interchange Format (AIF) \cite{chesnevar2006towards} was the product of an international effort to standardise the representation of argumentation, whilst having lossless translation into other formats. 

Argumentation has been applied in the legal domain, with Walton making a number of contributions \cite{walton2005argumentation,walton2010legal}. Argumentation has been demonstrated within the context of intelligence analysis in CISpaces \cite{toniolo2014argumentation}, a tool developed to support situational understanding in intelligence analysis.

In value based argumentation frameworks \cite{bench2002value} each argument is assigned a value. This introduces a notion of preference over arguments within the framework. This value is a completely different concept from the notion of value we consider in this paper. Bench-Capon considers argumentation frameworks with predefined values on arguments, whereas we look to determine a value from an argument framework. 

\subsection{Probabilistic Argumentation}

There have been a number of extensions of argumentation to uncertain arguments or attacks. \cite{hunter2017probabilistic} splits these approaches into two camps, the constellations approach and the epistemic approach. 

The constellations approach models uncertainty in the topology of the graph. Examples of constellation approaches are probabilistic argument frameworks \cite{li2011probabilistic}, AAJ \cite{dung2010towards} and using subgraphs in \cite{liao2018formulating}. 

Epistemic approaches model the uncertainty in the belief of an argument. Such approaches include probabilistic semantics \cite{thimm2012probabilistic}, dialogical uncertainties \cite{hunter2014probabilistic}, over incomplete data \cite{hunterthimm2014probabilistic} and equational semantics \cite{gabbay2015probabilistic}.

\subsection{Value of Information}\label{relatedvoi}

The concept of value of information was introduced by Howard \cite{howard1966information} in 1966 within the context of decision theory and has been the focus of significant research since. Much of the previous work on value of information has focussed on specific use cases, such as the oil industry \cite{bratvold2009value}, supply chain management \cite{quigley2018supplier} and the medical domain \cite{behrens2007learning}. The value of information has also been considered within abstract structures, such as influence diagrams \cite{dittmer2013myopic}, probabilistic graphic models \cite{krause2009optimal,ghosh2017optimal} and Probabilistic Logic Programs \cite{ghosh2019value}.

\section{Background}\label{background}

Many representations of argumentation have been developed, such as AIF \cite{chesnevar2006towards} and ASPIC+ \cite{prakken2010abstract}. Most abstract argumentation representations are underpinned by Dung frameworks, introduced in Dung's seminal paper \cite{dung1995acceptability} as a mathematical model to evaluate arguments. In this section we formally define Dung argument frameworks and a probabilistic extension to allow reasoning over uncertain arguments and attacks.

\subsection{Argumentation}

A Dung argumentation framework \cite{dung1995acceptability} is a model for arguments and attacks between these arguments. Arguments are represented as nodes in a graph, and attacks by directed edges from an argument to the argument it attacks. 

\begin{defn}[Dung Argumentation Framework]
A Dung Argumentation Framework (DAF) is a pair $F = (A,D)$ where $A$ is a set of arguments and $D \subseteq A\times A$ is a binary relation on attacks between arguments. $a$ is said to attack $b$ if $(a,b)\in D$.
\end{defn}

A set of arguments from an argument framework is known as an extension. Extensions are considered according to some semantics, which relate to conditions on the acceptability of arguments. Considering the extensions of an argument framework with respect to some semantics gives a set of extensions with some meaningful properties.  

\begin{defn}[Extension]
An extension of an argumentation framework $F = (A,D)$ is a subset of arguments $S \subseteq A$.
\end{defn}

\begin{defn}[Semantics for Argumentation Frameworks]
A semantics for argumentation frameworks is a function $\sigma$ assigning a set of extensions $\sigma(F) \subseteq 2^A$ for each argumentation framework $F = (A,D)$.
\end{defn}

Most semantics are defined on the basis of acceptability of arguments. An argument is accepted either if it is not attacked, or if it is defended by another argument, that is all of its attacking arguments are themselves attacked by other arguments.

\begin{defn}[Acceptable Argument]
Let $F = (A,D)$ be a DAF and $S \subseteq A$ an extension. Then an argument $a\in A$ is acceptable with respect to $S$ iff $\forall b\in A$ such that $(b,a)\in D$, $\exists c \in S$ such that $(c,b) \in D$.
\end{defn}

We can now define some common semantics on extensions. For a more extensive list of extension semantics, see \cite{charwat2015methods}. Let $F = (A,D)$ be an argumentation framework and $S \subseteq A$ an extension. Then:
\begin{itemize}
\item $S$ is conflict-free if there are no $a,b \in S$ such that $(a,b) \in D$.
\item $S$ is admissible if $S$ is conflict-free and $\forall a \in S$, $a$ is acceptable with respect to $S$.
\item $S$ is complete if $S$ is admissible and if $a\in A$ is acceptable with respect to $S$, then $a\in S$.
\item $S$ is grounded if $S$ is a minimal complete extension (under set inclusion).
\item $S$ is preferred if $S$ is a maximal complete extension (under set inclusion).
\end{itemize}

To evaluate a Dung argument framework, the extensions are calculated according to a semantics and inference is applied to each argument to determine if it is accepted. The inference modes most commonly used are credulous and sceptical acceptance. 

\begin{defn}[Credulous and Sceptical Acceptance]
Let $F = (A,D)$ be an argument framework, $\sigma$ a semantics and $a\in A$ an argument. $a$ is credulously accepted in $\sigma(F)$ if $a$ is accepted in some extension $S \in \sigma(F)$. $a$ is sceptically accepted in $\sigma(F)$ if $a$ is accepted in all extensions in $\sigma(F)$.
\end{defn}

\subsection{Probabilistic Argumentation}

Probabilistic argument frameworks \cite{li2011probabilistic,li2015probabilistic} are an extension of Dung argument frameworks, allowing numerical uncertainties against attacks and arguments.

\begin{defn}[Probabilistic Argumentation Framework]
A Probabilistic Argumentation Framework (PrAF) is a tuple $PF = (A, P_A, D, P_D)$ where $(A,D)$ is a Dung Argumentation Framework, $P_A \colon A \to (0,1]$ and $P_D \colon D \to (0,1]$ are probabilities on arguments and attacks respectively.
\end{defn}

An argument of probability zero would not exist in the framework, and therefore the probabilities are defined only on $(0,1]$. A PrAF builds a probabilistic representation of a DAF, with the aim of predicting the reasoning outputs of the underlying DAF. We call this underlying DAF the target DAF.

\begin{defn}[Target DAF]
Given PrAF $PF = (A, P_A, D, P_D)$, DAF $F^T = (A^T,D^T)$ with $A^T \subseteq A$ and $D^T \subseteq D \cap (A^T \times A^T)$ is the target DAF of PrAF iff for any argument $a\in A$, $P_A(a) = P(a\in A^T)$ and for any attack $(s,t)\in D$ such that $s,t \in A^T$, $P_D((s,t)) = P((s,t)\in D^T)$, where $P(a\in A^T)$ is the probability that the argument $a$ exists in $A^T$ and $P((s,t)\in D^T)$ is the probability that the attack $(s,t)$ exists in $D^T$, given that arguments $s$ and $t$ exist in $A^T$.
\end{defn}

Using the same semantics as for DAFs, PrAFs can be evaluated to determine the probability that an extension is an extension of the target DAF according to the semantics. This is known as the probabilistic justification of an extension. 

\begin{defn}[Probabilistic Justification]
Given a PrAF $PF = (A, P_A, D, P_D)$, its target DAF $F^T = (A^T,D^T)$, DAF semantics $\sigma$ and an extension $S \subseteq A$, the probabilistic justification of $S$ with respect to semantics $\sigma$, $P_{PF}^{\sigma}(S)$ is the probability that $S \in \sigma(F^T)$.
\end{defn}

The notion of probabilistic justification provides a predicted probability on whether arguments are justified in the target DAF, but as the target DAF cannot be fully determined, we cannot compute it. Instead we must consider the space of all possible target DAFs, known as inducible DAFs.

\begin{defn}[Inducible DAF]
A DAF $F = (A',D')$ is induced from a probabilistic argumentation framework $PF = (A, P_A, D, P_D)$ iff all of the following hold:
\begin{itemize}
\item $A' \subseteq A$.
\item $D' \subseteq D \cap (A'\times A')$.
\item $\forall a \in A$ such that $P_A(a)=1$, $a \in A'$.
\item $\forall (f,t) \in D$ such that $P_D((f,t))=1$ and $P_A(f) = P_A(t) = 1$, $(f,t)\in D'$.
\end{itemize}
\end{defn}

To simplify the calculations, we assume that the existence in the target DAF of an argument is independent from the existence of any other argument, and the existence of an attack is independent of the existence of any other attack. Approaches to relax this assumption are detailed in \cite{li2015probabilistic}. This assumption allows the calculation of the probability of an inducible DAF $F^I = (A^I,D^I)$ by 
\begin{multline*}
P_{PF}^I(F^I) = \\ \prod_{a \in A^I} P_A(a) \prod_{a \in A \setminus A^I} (1-P_A(a)) \prod_{d \in D^I} P_D(d) \prod_{d \in (D \cap (A^I \times A^I)) \setminus D^I} (1-P_D(d)).
\end{multline*}

A probabilistic justification for each extension can be calculated by iterating through all inducible DAFs. The same credible and sceptical inference modes can be applied as for DAFs to provide a probability against each argument. Due to the exponential complexity of iterating though all inducible DAFs, for large graphs the probability is approximated by Monte Carlo methods.

\section{Value of Information in Argumentation}\label{VoI}

Understanding the value of information within an argument framework allows the identification of the arguments which underpin the result of an evaluation of the framework. A value of information could also help identify the most important nodes in an argument framework to attack, in order to change the outcome of an evaluation. In legal argumentation, this could identify the arguments which the Prosecution and Defence should focus on to make their case. For argumentation in intelligence analysis, this will help to identify intelligence gaps, and help to direct the collection of further evidence to support or refute the hypotheses being evaluated. 

The value of an item of information is inherently dependent on the available evidence and the question being answered. The argument framework provides a formal representation of the available evidence. Within an argument map there are a number of questions that could be asked. The objective could be to make the arguments in an extension acceptable, or to effect a change on the result of an evaluation. We formally define the objective for an (Dung or probabilistic) argument framework $F$ as an extension $O\subseteq A$, where $A$ is the set of arguments in $F$, and a utility function $U \colon O \times \mathcal{F} \to \mathbb{R}$, where $\mathcal{F}$ is the set of all (Dung or probabilistic) argument frameworks.

As described in section \ref{relatedvoi}, a number of different constructs have been used for the value of information. Common objectives are the minimisation of uncertainty in the outputs, minimising entropy, and maximising information gain. 

Within an argument framework, we can consider the value of arguments currently in the framework, but we can also consider the value of adding a new argument to the framework. We define these as the  value of observed and value of observation respectively. 

\begin{defn}[Value of Observed]
Let $F$ be an argument framework, $O\subseteq A$ the objective extension, where $A$ is the set of arguments in $F$, and $U \colon O \times \mathcal{F} \to \mathbb{R}$ a utility function. Then the value for observed is a function $V_{observed} \colon 2^A \to \mathbb{R}$, defined by 
$$V_{observed}(\alpha) = \sum_{e\in O} d(U(e,F),U(e,F\setminus \{\alpha\})),$$ 
where $\alpha \subset A$ and $F\setminus \{\alpha\}$ is the framework $F$ with the arguments $\alpha$ and all associated attacks removed, and $d\colon \mathbb{R}^2 \to \mathbb{R}$ is a function representing the difference we want to maximise.
\end{defn}

\begin{defn}[Value of Observation]
Let $F$ be an argument framework, $O\subseteq A$ the objective extension, where $A$ is the set of arguments in $F$, and $U \colon O \times \mathcal{F} \to \mathbb{R}$ a utility function. Then the value for observation is a real function $V_{observation}$ defined on tuples $B$ of arguments $\alpha$ such that $\alpha \cap A = \emptyset$ and attacks $(a,b) \in \alpha \times A \cup A \times \alpha \cup \alpha \times \alpha$, with associated probabilities if $F$ is a probabilistic argumentation framework. $V_{observation}$ is defined by 
$$V_{observation}(B) = \sum_{e\in O}d(U(e,F \cup B),U(e,F)),$$ where $F \cup B$ is the argument framework $F$ with the arguments and attacks in $B$ added, and $d\colon \mathbb{R}^2 \to \mathbb{R}$ is a function representing the difference we want to maximise.
\end{defn}

The value of observed is defined on the finite closed world of arguments currently in the argument framework, whereas the value of an observation is defined on the open world of possible arguments and attacks which could be added to the argument framework. 

\subsection{Argumentation}

The binary output of the evaluation of a Dung framework limits the utility functions supporting a value function. Let $F = (A,D)$ be an argument framework and $O\subseteq A$ the objective extension. Two examples of utility functions that can be used for DAFs are:
\begin{itemize}
\item \textbf{Target output}: If there is a target state of the objective, with accepted arguments $O' \subseteq O$, a utility function can be defined by evaluating the Dung framework $F$ with respect to some semantics and inference mode, which will result in a set of accepted arguments $E$. Then 
$$U(e,F) = \begin{cases} 1 &\text{if } e \in (O'\cap E) \cup (O\setminus O' \cap O \setminus E) \\ 0 &\text{otherwise.} \end{cases}$$
With $d(x,y) = x-y$, an argument will be valuable if it moves the set of accepted arguments towards the desired state. 
\item \textbf{Maximising change}: A utility function can be defined by evaluating the Dung framework $F$ with respect to some semantics and inference mode, which will result in a set of accepted arguments $E$. Then 
$$U(e,F) = \begin{cases} 1 &\text{if } e \in E \\ 0 &\text{otherwise.} \end{cases}$$ 
With $d(x,y) = |x-y|$, an argument will be valuable if it changes the evaluation of the Dung framework.
\end{itemize}

\begin{figure}[htb]
	\centering
		\includegraphics[scale=0.7]{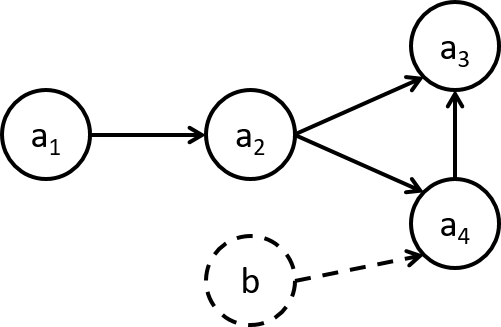}
	\caption{The Dung argument framework from example \ref{daf}.}
	\label{fig:DAFExample}
\end{figure}

\begin{example}\label{daf}
Let $F = (A,D)$ be a Dung argument framework with arguments $A = \{a_1,a_2,a_3,a_4\}$ and attacks $D = \{(a_1,a_2),(a_2,a_3),(a_2,a_4),(a_4,a_3)\}$, as shown in figure \ref{fig:DAFExample}. Then $F$ under grounded semantics has only one extension $E = \{a_1,a_4\}$, and so the arguments $a_1$ and $a_4$ are sceptically accepted. In this example, take objective to be defined by extension $O = \{a_3,a_4\}$ and utility function defined as above to target the output $O' = \{a_3\}$. By removing argument $a_1$ and its related attacks from the framework, the sceptically accepted arguments are $\{a_2\}$ Therefore, with $d(x,y)=x-y$, $V_{observed}(\{a_1\}) = (0-0) + (0-1) = -1$, so the argument $a_1$ is detracting from the desired output. Now consider adding the argument $b$, with attack $\{(b,a_4)\}$. Then the sceptically accepted arguments are $\{a_1,a_3,b\}$ and $V_{observation}((\{b\},\{(b,a_4)\})) = (1-0)+(1-0) = 2$. Therefore attacking the argument $a_4$ would be very valuable if the objective is to output $\{a_3\}$. 
\end{example}

The target output utility function introduces confirmation bias, so would not be suitable for the intelligence analysis use case.

\subsection{Probabilistic Argumentation}

Probabilistic argumentation frameworks output a probability against each argument after evaluation which allows a much richer theory of value. Let $PF = (A,P_A,D,P_D)$ be a probabilistic argument framework and $O\subseteq A$ the objective extension. \cite{krause2009optimal} presents a wide range of utility functions for information value. Here we select a few:
\begin{itemize}
\item \textbf{Target output}: If there is a target set of objective arguments to be accepted $O' \subseteq O$, a utility function can be defined by evaluating the PrAF $PF$ with respect to some semantics and inference mode, which will result in a set probability assigned to each argument $P\colon A \to [0,1]$. Then 
$$U(e,PF) = \begin{cases} P(e) &\text{if } e \in O' \\ 1-P(e) &\text{otherwise.} \end{cases}$$ 
With $d(x,y) = x-y$, an argument will be valuable if it moves the set of accepted arguments towards the desired state. 
\item \textbf{Minimising entropy}: A utility function can be defined by evaluating the PrAF $PF$ with respect to some semantics and inference mode, which will result in a set probability assigned to each argument $P\colon A \to [0,1]$. Then $U(e,PF) = -H(P(e)) = P(e)\log P(e) + (1-P(e))\log(1-P(e))$. With $d(x,y) = x-y$, an argument will be valuable if it minimises the entropy of PrAF output on the desired extension. 
\item \textbf{Maximising change}: A utility function can be defined by evaluating the PrAF $PF$ with respect to some semantics and inference mode, which will result in a set probability assigned to each argument $P\colon A \to [0,1]$. Then $U(e,PF) = P(e)$. With $d(x,y) = |x-y|$, an argument will be valuable if it changes the evaluation of the PrAF.
\item \textbf{Kullback–Leibler (KL) Divergence} A utility function can be defined by evaluating the PrAF $PF$ with respect to some semantics and inference mode, which will result in a set probability assigned to each argument $P\colon A \to [0,1]$. Then $U(e,PF) = P(e)$. With $d(x,y) = KL([x,1-x]\parallel [y,1-y]) = x\log\left(\frac{x}{y}\right) + (1-x)\log\left(\frac{1-x}{1-y}\right)$. This is an information theoretic way to construct a value function in which an argument is valuable if it changes the evaluation of the PrAF.
\end{itemize}

\begin{figure}[htb]
	\centering
		\includegraphics[scale=0.7]{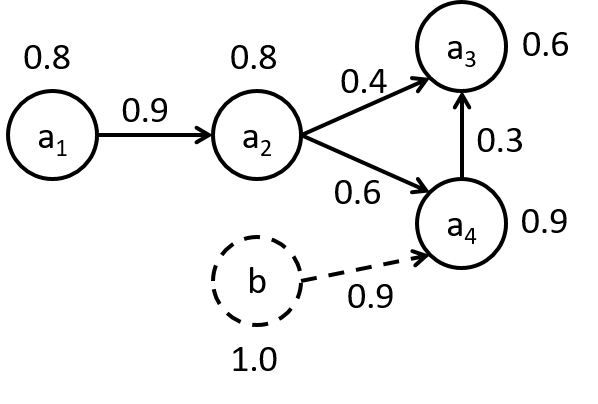}
	\caption{The probabilistic argument framework from example \ref{praf}.}
	\label{fig:PrAFExample}
\end{figure}

\begin{example}\label{praf}
Let $PF = (A,P_A,D,P_D)$ be a probabilistic argument framework with arguments $A = \{a_1, a_2, a_3, a_4\}$, attacks $D = \{(a_1,a_2), (a_2,a_3), (a_2,a_4), (a_4,a_3)\}$ and probabilities $P_A$, $P_D$ on $A$ and $D$ respectively as shown in figure \ref{fig:PrAFExample}. Then under grounded semantics and sceptical inference the probabilities for each argument in $A$ are $P(a_1)=0.8000$, $P(a_2)=0.2240$, $P(a_3)=0.4118$ and $P(a_4)=0.7790$. As above, define an objective on $PF$ by extension $O = \{a_3,a_4\}$ and utility function defined to target the output $\{a_3\}$. By removing argument $a_1$ and its related attacks from the framework, the probabilities for each argument are $P(a_2)=0.8000$, $P(a_3)=0.3445$ and $P(a_4)=0.4680$. Therefore 
$$V_{observed}(\{a_1\}) = (0.4118-0.3445)+((1-0.7790)-(1-0.4680)) = -0.2435.$$
Now consider adding the argument $b$ with probability $P_A^b(b) = 1$, and attack $\{(b,a_4)\}$ with probability $P_D^b((b,a_3)) = 0.9$. Then the argument probabilities are $P(a_1)=0.8000$, $P(a_2)=0.2240$, $P(a_3)=0.5328$, $P(a_4)=0.0779$ and $P(b)=1.000$. Therefore 
\begin{multline*}
V_{observation}((\{b\},P_A^b,\{(b,a_4)\},P_D^b)) = (0.5328-0.4118)- \\((1-0.0779)-(1-0.7790) = 0.8221.
\end{multline*}
\end{example}

\section{Argumentation for Intelligence Analysis}\label{ApplicationIA}

In an intelligence analysis, analysts assess the evidence available to them, and perform inference on this evidence to support or refute the hypotheses they are assessing. Argumentation theory provides a mathematical model which can represent the arguments made during an intelligence analysis. Confirmation bias is a key consideration when performing an intelligence analysis. In order to reduce confirmation bias, an item of evidence should be valuable if it changes the output of the reasoning, rather than confirming or refuting a given hypothesis. 

In order to reduce the biases in an intelligence analysis, analysts use structured analytical techniques \cite{uk2013quick} such as Analysis of Competing Hypotheses (ACH) \cite{heuer1999psychology}. The ACH structured analytical technique is underpinned by identifying inconsistencies between the evidence available and the hypotheses being assessed. This is generally represented in a matrix, with hypotheses along one axis and evidence along the other. For each hypothesis and evidence pair, a label is added in the ACH matrix to represent consistency, inconsistency or not applicable. ACH can be represented in argumentation by an attack between the arguments representing evidence and hypotheses if they are inconsistent. This representation of ACH as an argument and the value of information functions defined in this paper means we can now consider the value of evidence in an ACH. Using argumentation to represent an intelligence analysis can reduce confirmation bias in the same way as ACH can, as the focus is on finding conflicts between evidence and hypotheses, rather than evidence to support hypotheses.

Analysts may also attach uncertainties to evidence, based on the confidence from the intelligence source. We can therefore represent intelligence analyses as both Dung frameworks and PrAFs, and apply the values of information constructed above. In the evaluation of PrAFs, we assume the evidence is independent. This may not be true in real intelligence analyses. In his thesis \cite{li2015probabilistic}, Li proposes probabilistic evidential argumentation frameworks to relax this independence assumption. 

The objective of an intelligence analysis is to confirm or deny a set of hypotheses. Therefore an objective should be defined with respect to an extension $E$ consisting of arguments which correspond to hypotheses in the analysis.  In order to minimise confirmation bias, and avoid the value function reinforcing the most likely hypotheses, the utility function must be carefully chosen. In particular we define information to be valuable if it changes the output of the analysis. This means that both evidence to support and disprove the hypotheses will be valuable. We will therefore use the maximising change utility functions defined above for DAFs and the KL divergence as an information theoretic approach for PrAFs.

\subsection{Example}

To demonstrate argumentation and value of information in an intelligence context, we provide an example in the maritime domain. On 16\textsuperscript{th} August 2020, the chemical tanker Aegean II was allegedly hijacked by armed men in Somalia waters, while en route from UAE to Mogadishu, Somalia. Whilst travelling slowly through Somalian waters, with an apparent mechanical problem, a sudden change of direction was observed in AIS tracking. Every ship has Automated Identification System (AIS) transceiver which transmits information, including location, that is used by vessel traffic services. It has been claimed that the Aegean II was attacked by six armed men. There were conflicting reports on whether the Aegean II was hijacked by pirates, or seized by a local militia, who act as the local police. There is a history of piracy in Somalian waters, although there have been no piracy incidents in the preceding 3 years. In the following example, we produce a simple ACH analysis of these two hypotheses on who hijacked the Aegean II, constructed purely for demonstration, using only open source news reporting.

Table \ref{tab:AIIACH} shows the ACH matrix for this problem. Each hypothesis and evidence pair is labelled with \textit{I} or \textit{II} if they are weakly inconsistent or inconsistent respectively, \textit{C} or \textit{CC} if they are weakly consistent or consistent respectively or \textit{N/A} if the evidence is neither consistent or inconsistent with the hypothesis. We are \textit{certain} on evidence $e_2$, but there is some uncertainty around evidence $e_1$ and $e_3$ so we label it \textit{likely}. 

\begin{table}[h]
	\centering
		\begin{tabular}{|M{4cm}|M{3cm}|M{3cm}|}
		  \cline{2-3}
      \multicolumn{1}{c|}{} & $h_1$: The Aegean II was hijacked by pirates & $h_2$: The Aegean II was seized by local police/media \\ \hline
			$e_1$: The men boarding the Aegean II were wearing police uniforms (\textit{likely}) & II & C \\ \hline
			$e_2$: History of piracy in Somalian waters (\textit{certain}) & C & I \\ \hline
			$e_3$: Local media reports the Aegean II was seized by local police/militia (\textit{likely}) & II & CC \\ \hline
		\end{tabular}
		\caption{An analysis of competing hypotheses (ACH) on the hijacking of Aegean II.}
		\label{tab:AIIACH}
\end{table}

To convert the ACH into a Dung argument framework, we take the arguments to be the evidence and hypotheses, and attacks from an item of evidence to a hypothesis if they are labelled as inconsistent (\textit{I} or \textit{II}) in the ACH matrix. The hypotheses are assumed to be mutually exclusive so we also add attacks between the two hypotheses. 

\begin{figure}[htb]
	\centering
		\includegraphics[scale=0.7]{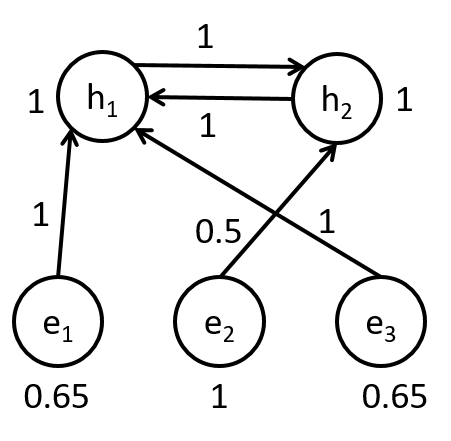}
	\caption{The ACH in table \ref{tab:AIIACH}, represented as a probabilistic argument framework.}
	\label{fig:AIIArgument}
\end{figure}

The ACH can also be viewed as a probabilistic argument framework by mapping the uncertainties and strong and weak inconsistencies to numerical probabilities. For uncertainties we define certain to be probability $1$ and likely as probability $0.65$, and for inconsistencies, weakly inconsistent (\textit{I}) is an attack with probability $0.5$ and inconsistent (\textit{II}) is an attack with probability $1$. These probabilities were selected arbitrarily for this example. For the hypotheses, which we are testing in this analysis, we assign probability $1$. The hypotheses and attacks between hypotheses are assigned probability $1$, as this is what we are testing. A diagram of the PrAF produced is shown in figure \ref{fig:AIIArgument}.

Evaluating this as a Dung framework, with grounded semantics and sceptical inference, the arguments $e_1,e_2,e_3$ are accepted. Therefore given the information in the ACH we have evidence against both hypotheses. Taking objective defined by the extension containing the hypotheses $E = \{h_1,h_2\}$, with maximising change utility function and $d(x,y)=|x-y|$, we have 
\begin{multline*}
V_{observed}(\{e_2\}) = V_{observed}(\{e_1,e_2\}) = V_{observed}(\{e_1,e_3\}) \\ = V_{observed}(\{e_2,e_3\}) = 1
\end{multline*}
and $V_{observed}(S) = 0$ otherwise, for $S \subseteq \{e_1,e_2,e_3\}$. For the value of observation, we compare attacking $h_2$ and $e_3$. Attacking $h_2$ with some argument $b$ we have $V_{observation}((\{b\},\{(b,h_2)\})) = 1$, whilst attacking $e_3$ with some argument $b$ we have $V_{observation}((\{b\},\{(b,e_3)\})) = 0$. Therefore evidence to refute hypothesis $h_2$ is more valuable to the analysis as evidence to refute evidence $e_3$.

As a probabilistic argumentation framework, this can be evaluated to give probabilities on arguments $P(e_1) = 0.65$, $P(e_2) = 1$, $P(e_3) = 0.65$, $P(h_1) = 0.06125$ and $P(h_2) = 0.43875$. Again we see hypothesis $h_1$ is rejected but we also have a degree of belief on hypothesis $h_2$. With objective defined by the extension containing the hypotheses $E = \{h_1,h_2\}$, utility function $U(e,F) = P(e)$ and $d$ the KL-divergence (calculated with base $e$), we calculate $V_{observed}(\{e_1\}) = V_{observed}(\{e_3\}) = 0.0850$ and $V_{observed}(S)$ is undefined for any other $S \subseteq \{e_1,e_2,e_3\}$, but we consider it as $+\infty$. This tells us the analysis is not particularly well founded, as it is easy to significantly change the output by removing observations. If we compare adding an uncertain argument (with probability $0.5$), attacking (with probability $1$) $h_2$ and $e_3$ respectively, we see that 
$$V_{observation}((\{b\},P_A^b,\{(b,h_2)\},P_D^b)) = 0.1126$$
and
$$V_{observation}((\{b\},P_A^b,\{(b,e_3)\},P_D^b)) = 0.0291.$$
Therefore uncertain evidence which refutes the hypothesis $h_2$ is more valuable than attacking the evidence $e_3$. Using PrAF, we also have a figure of how much more valuable attacking $h_2$ is than $e_3$.

The values of these observations can then be balanced against the risks and costs of the potential methods of collection, such as sending a team to approach the Aegean II or sending a team to investigate at the port, to ensure the most suitable collection opportunity is taken.

The Aegean II was released safely on the 21\textsuperscript{st} August 2020. The EU NAVFOR counter piracy operation cited a Bossaso port authority reporting that the Aegean II had not been hijacked and that Garadfoo police had been sent on board to inspect the vessel when she was drifting off Bereeda waiting for technical assistance.

\section{Conclusions}\label{conclusions}

This paper has introduced a notion of value of information within argument frameworks. We have introduced a family of functions representing the value of arguments existing in an argument framework, and also for arguments and attacks which could be added to an argument framework. These value functions could help identify the important arguments in a framework, and the attacks which could change the evaluation of an argument framework, to suit our objective. This could be particularly useful in legal argumentation - to identify the areas the Prosecution and Defence should focus on; and Intelligence analysis - to inform the collection of further evidence. We have demonstrated how argument frameworks could be used in the latter application in a maritime piracy example.

\section{Future Work}\label{futurework}

In this paper, only argument frameworks with simple one-to-one attacks have been considered. Argument Interchange Format (AIF) \cite{chesnevar2006towards} allows the representation of many-to-one conflicts, but also allows the representation of evidence to support arguments. AIF does not provide a method for evaluation and the Dung argument framework does not represent these complex arguments. ASPIC+ \cite{prakken2010abstract} provides a way to transform arguments in AIF into Dung frameworks \cite{bex2010formal}, and therefore enable the value functions defined in this paper to be applied to a wider range of arguments. For uncertain arguments, with conflict and evidential relationships, probabilistic evidential argumentation frameworks \cite{li2015probabilistic} may provide a method to extend the value of information to arguments represented in AIF. In intelligence analysis, this would allow the representation and application of the value function to more complex arguments. 

Whilst understanding the value of information in an argument map is useful, when considering the value of observations this must be balanced by the cost of making the observation. In intelligence analysis the collection of further intelligence can be requested to improve the confidence in an analysis. The collection of intelligence can be expensive and may pose a risk to assets. It is therefore often not possible to collect all the evidence required to make a conclusion with full confidence. The value of making a collection to improve the confidence in the analysis outputs must be balanced against the cost and risk of making the collection. The value functions defined in this paper, when balanced against a cost metric, could help to prioritise the collection of intelligence and understand when an analysis has sufficient evidence.

\section*{References}

\bibliography{VoIArgumentation}

\begin{thebibliography}{10}
\expandafter\ifx\csname url\endcsname\relax
  \def\url#1{\texttt{#1}}\fi
\expandafter\ifx\csname urlprefix\endcsname\relax\def\urlprefix{URL }\fi
\expandafter\ifx\csname href\endcsname\relax
  \def\href#1#2{#2} \def\path#1{#1}\fi

\bibitem{dung1995acceptability}
P.~M. Dung, On the acceptability of arguments and its fundamental role in
  nonmonotonic reasoning, logic programming and n-person games, Artificial
  intelligence 77~(2) (1995) 321--357.

\bibitem{uk2013quick}
{UK Ministry of Defence}, Quick wins for busy analysts (2013).

\bibitem{murukannaiah2015resolving}
P.~K. Murukannaiah, A.~K. Kalia, P.~R. Telangy, M.~P. Singh, Resolving goal
  conflicts via argumentation-based analysis of competing hypotheses, in: 2015
  IEEE 23rd International Requirements Engineering Conference (RE), IEEE, 2015,
  pp. 156--165.

\bibitem{nickerson1998confirmation}
R.~S. Nickerson, Confirmation bias: A ubiquitous phenomenon in many guises,
  Review of general psychology 2~(2) (1998) 175--220.

\bibitem{prakken2010abstract}
H.~Prakken, An abstract framework for argumentation with structured arguments,
  Argument and Computation 1~(2) (2010) 93--124.

\bibitem{walton2009argumentation}
D.~Walton, Argumentation theory: A very short introduction, in: Argumentation
  in artificial intelligence, Springer, 2009, pp. 1--22.

\bibitem{baroni2011introduction}
P.~Baroni, M.~Caminada, M.~Giacomin, An introduction to argumentation
  semantics, Knowledge Engineering Review 26~(4) (2011) 365.

\bibitem{rahwan2011representing}
I.~Rahwan, B.~Banihashemi, C.~Reed, D.~Walton, S.~Abdallah, Representing and
  classifying arguments on the semantic web, Knowledge Engineering Review
  26~(4) (2011) 487--511.

\bibitem{ter2013critical}
T.~ter Berg, T.~Van~Gelder, F.~Patterson, S.~Teppema, Critical thinking:
  Reasoning and communicating with rationale, Critical Thinking Skills, 2013.

\bibitem{chesnevar2006towards}
C.~Chesnevar, S.~Modgil, I.~Rahwan, C.~Reed, G.~Simari, M.~South, G.~Vreeswijk,
  S.~Willmott, et~al., Towards an argument interchange format, The knowledge
  engineering review 21~(4) (2006) 293--316.

\bibitem{walton2005argumentation}
D.~Walton, Argumentation methods for artificial intelligence in law, Springer
  Science \& Business Media, 2005.

\bibitem{walton2010legal}
D.~Walton, Legal argumentation and evidence, Penn State Press, 2010.

\bibitem{toniolo2014argumentation}
A.~Toniolo, T.~Dropps, W.~R. Ouyang, J.~A. Allen, T.~J. Norman, N.~Oren, M.~B.
  Srivastava, P.~Sullivan, Argumentation-based collaborative intelligence
  analysis in cispaces., in: COMMA, 2014, pp. 481--482.

\bibitem{bench2002value}
T.~Bench-Capon, Value based argumentation frameworks, arXiv preprint
  cs/0207059.

\bibitem{hunter2017probabilistic}
A.~Hunter, M.~Thimm, Probabilistic reasoning with abstract argumentation
  frameworks, Journal of Artificial Intelligence Research 59 (2017) 565--611.

\bibitem{li2011probabilistic}
H.~Li, N.~Oren, T.~J. Norman, Probabilistic argumentation frameworks, in:
  International Workshop on Theorie and Applications of Formal Argumentation,
  Springer, 2011, pp. 1--16.

\bibitem{dung2010towards}
P.~M. Dung, P.~M. Thang, Towards (probabilistic) argumentation for jury-based
  dispute resolution., COMMA 216 (2010) 171--182.

\bibitem{liao2018formulating}
B.~Liao, K.~Xu, H.~Huang, Formulating semantics of probabilistic argumentation
  by characterizing subgraphs: theory and empirical results, Journal of Logic
  and Computation 28~(2) (2018) 305--335.

\bibitem{thimm2012probabilistic}
M.~Thimm, A probabilistic semantics for abstract argumentation., in: ECAI,
  Vol.~12, 2012, pp. 750--755.

\bibitem{hunter2014probabilistic}
A.~Hunter, Probabilistic strategies in dialogical argumentation, in:
  International Conference on Scalable Uncertainty Management, Springer, 2014,
  pp. 190--202.

\bibitem{hunterthimm2014probabilistic}
A.~Hunter, M.~Thimm, Probabilistic argumentation with incomplete information.,
  in: ECAI, 2014, pp. 1033--1034.

\bibitem{gabbay2015probabilistic}
D.~M. Gabbay, O.~Rodrigues, Probabilistic argumentation: An equational
  approach, Logica Universalis 9~(3) (2015) 345--382.

\bibitem{howard1966information}
R.~A. Howard, Information value theory, IEEE Transactions on systems science
  and cybernetics 2~(1) (1966) 22--26.

\bibitem{bratvold2009value}
R.~B. Bratvold, J.~E. Bickel, H.~P. Lohne, et~al., Value of information in the
  oil and gas industry: past, present, and future, SPE Reservoir Evaluation \&
  Engineering 12~(04) (2009) 630--638.

\bibitem{quigley2018supplier}
J.~Quigley, L.~Walls, G.~Demirel, B.~L. MacCarthy, M.~Parsa, Supplier quality
  improvement: The value of information under uncertainty, European journal of
  operational research 264~(3) (2018) 932--947.

\bibitem{behrens2007learning}
T.~E. Behrens, M.~W. Woolrich, M.~E. Walton, M.~F. Rushworth, Learning the
  value of information in an uncertain world, Nature neuroscience 10~(9) (2007)
  1214--1221.

\bibitem{dittmer2013myopic}
S.~L. Dittmer, F.~V. Jensen, Myopic value of information in influence diagrams,
  arXiv preprint arXiv:1302.1535.

\bibitem{krause2009optimal}
A.~Krause, C.~Guestrin, Optimal value of information in graphical models,
  Journal of Artificial Intelligence Research 35 (2009) 557--591.

\bibitem{ghosh2017optimal}
S.~Ghosh, C.~Ramakrishnan, Optimal value of information in dynamic bayesian
  networks, in: 2017 IEEE 29th International Conference on Tools with
  Artificial Intelligence (ICTAI), IEEE, 2017, pp. 16--23.

\bibitem{ghosh2019value}
S.~Ghosh, C.~Ramakrishnan, Value of information in probabilistic logic
  programs, arXiv preprint arXiv:1909.08234.

\bibitem{charwat2015methods}
G.~Charwat, W.~Dvo{\v{r}}{\'a}k, S.~A. Gaggl, J.~P. Wallner, S.~Woltran,
  Methods for solving reasoning problems in abstract argumentation--a survey,
  Artificial intelligence 220 (2015) 28--63.

\bibitem{li2015probabilistic}
H.~Li, Probabilistic argumentation, Ph.D. thesis, Citeseer (2015).

\bibitem{heuer1999psychology}
R.~J. Heuer, Psychology of intelligence analysis, Center for the Study of
  Intelligence, 1999.

\bibitem{bex2010formal}
F.~Bex, H.~Prakken, C.~Reed, A formal analysis of the aif in terms of the aspic
  framework., in: COMMA, 2010, pp. 99--110.

\end{thebibliography}

\end{document}